\documentclass[
]{ceurart}

\usepackage{listings}
\usepackage{subfig}
\usepackage[numbers]{natbib}
\usepackage{placeins}

\begin{document}

\copyrightyear{2025}
\copyrightclause{Copyright for this paper by its authors.
Use permitted under Creative Commons License Attribution 4.0
International (CC BY 4.0).}


\conference{{D}e{F}actify 4.0: Fourth workshop on Multimodal Fact-Checking and Hate Speech Detection, March 2025, Philadelphia, Pennsylvania, USA}

\title{A Comprehensive Dataset for Human vs. AI Generated Text Detection}

\tnotemark[1]
\tnotetext[1]{This document is based on the CEUR-WS template and incorporates topics inspired by the Defactify workshop series.}

\author[1]{Rajarshi Roy}[email=royrajarshi0123@gmail.com]
\author[2]{Gurpreet Singh}
\author[3]{Ashhar Aziz}
\author[4]{Shashwat Bajpai}
\author[5]{Nasrin Imanpour}
\author[6]{Shwetangshu Biswas}
\author[7]{Kapil Wanaskar}
\author[8]{Parth Patwa}
\author[9]{Subhankar Ghosh}
\author[10]{Shreyas Dixit}
\author[1]{Nilesh Ranjan Pal}
\author[5]{Vipula Rawte}
\author[5]{Ritvik Garimella}
\author[11]{Gaytri Jena}
\author[12]{Amitava Das}
\author[5]{Amit Sheth}
\author[13]{Vasu Sharma}
\author[14]{Aishwarya Naresh Reganti}
\author[13]{Vinija Jain}
\author[14]{Aman Chadha}
\address{$^1$Kalyani Government Engineering College, India. $^2$IIIT Guwahati, India. $^3$IIIT Delhi, India. $^4$BITS Pilani Hyderabad Campus, India. $^5$University of South Carolina, USA. $^6$NIT Silchar, India. $^7$San José State University, USA. $^8$UCLA, USA. $^9$Washington State University, USA. $^{10}$Vishwakarma Institute of Information Technology, India. $^{11}$Gandhi Institute for Technological Advancement, India. $^{12}$BITS Pilani Goa, India. $^{13}$Meta AI, USA. $^{14}$Amazon AI, USA.}

\begin{abstract}
The rapid advancement of large language models (LLMs) has led to increasingly human-like AI-generated text, raising concerns about content authenticity, misinformation, and trustworthiness. Addressing the challenge of reliably detecting AI-generated text and attributing it to specific models requires large-scale, diverse, and well-annotated datasets. In this work, we present a comprehensive dataset comprising over 73,000 text samples that combine authentic New York Times articles with synthetic versions generated by multiple state-of-the-art LLMs including Gemma-2-9B, Mistral-7B, Qwen-2-72B, LLaMA-8B, Yi-Large, and GPT-4o. We establish baseline results for two key tasks: distinguishing human-written from AI-generated text, achieving an accuracy of 53\%, and attributing AI texts to their generating models with an accuracy of 5.04\%. By bridging real-world journalistic content with modern generative models, the dataset aims to catalyze the development of robust detection and attribution methods, fostering trust and transparency in the era of generative AI. Our dataset is available at: https://huggingface.co/datasets/Rajarshi-Roy-research/Defactify\_Text\_Dataset.
\end{abstract}


\begin{keywords}
AI-generated text detection \sep large language models \sep dataset \sep text attribution \sep misinformation
\end{keywords}

\maketitle

\section{Introduction}

The recent surge in the capabilities of large language models (LLMs) has dramatically transformed the landscape of text generation. State-of-the-art AI systems can now produce highly coherent and contextually relevant narratives, reports, and articles that are increasingly difficult to distinguish from human-written content. While these advances unlock significant opportunities for innovation across domains such as journalism, education, and digital assistants, they also intensify concerns regarding authenticity, misinformation, and fake news. The problem of online fake news, especially related to topics like pandemic, elections \cite{patwa2021overview,iranfakenews,morales2021covid19testsgonerogue}, etc. is exacerbated by AI generated fake news.

Amidst this evolving environment, the challenge of reliably detecting and attributing AI-generated text has become a research priority \cite{zellers2020defendingneuralfakenews,mitchell2023detectgptzeroshotmachinegeneratedtext}. The capability to differentiate between human-authored and machine-generated content is crucial for safeguarding the integrity of news media, academic publishing, and online discourse. However, progress in this area has been hindered by the scarcity of large, diverse, and well-annotated datasets that contain both human and AI-generated texts under realistic, real-world conditions.

In this paper, we introduce a dataset to detect AI generated text. The dataset is a comprehensive, extensively annotated resource curated to advance research on the detection of AI-generated content. Our dataset is built upon a large-scale collection of authentic news articles from the New York Times, spanning over two decades and covering a broad spectrum of topics. For each sampled article, we provide both the original human-authored narrative and multiple synthetic counterparts generated by leading LLMs, including Gemma-2-9B \cite{gemmateam2024gemma2improvingopen}, Mistral-7B \cite{jiang2023mistral7b}, Qwen-2-72B \cite{bai2023qwentechnicalreport}, LLaMA-8B  \cite{touvron2023llamaopenefficientfoundation}, Yi-Large \cite{ai2025yiopenfoundationmodels}, and GPT-4o \cite{openai2024gpt4o}. Each sample is enriched with detailed metadata, facilitating nuanced analysis and enabling a wide range of research applications.

The unique structure of the dataset supports tasks such as binary classification (distinguishing human from AI content) and model attribution (identifying the specific LLM responsible for a synthetic text). By bridging the gap between real-world journalistic data and diverse state-of-the-art generative models, our work lays the foundation for more robust benchmarks, and tools that are urgently needed to maintain trust, clarity, and accountability in the age of generative AI.

\section{Related Work}

 Recently, there has been significant research in two interconnected domains: detecting AI-generated text and identifying fake news. Both areas require large-scale, diverse, and well-annotated datasets to develop robust detection methods. In this section, we survey existing datasets and methods in these two critical research areas.

\subsection{AI-Generated Text Detection}

The development of effective AI-generated text detectors relies heavily on high-quality benchmark datasets. Several recent efforts have contributed valuable resources to this domain.

\textbf{RAID} \cite{raid2024} represents the largest and most challenging benchmark for machine-generated text detection, containing over 6 million text generations spanning 11 different language models across 8 diverse domains.
\textbf{M4} \cite{m42024} is a multi-generator, multi-domain, and multi-lingual dataset that includes text generated by multiple LLMs across seven languages, coverage of low-resource languages such as Arabic, Urdu, Indonesian, and Bulgarian. 
\textbf{TuringBench} \cite{uchendu2021turingbenchbenchmarkenvironmentturing} provides a benchmark environment containing approximately 10,000 news articles, predominantly focused on political content, with synthetic counterparts generated by 19 different language models including various GPT-2 variants \cite{radford2019language}, GPT-3 \cite{brown2020language}, GROVER models \cite{zellers2019defending}, CTRL \cite{keskar2019ctrl}, XLM \cite{lample2019cross}, XLNET \cite{yang2019xlnet}, and Transformer-XL \cite{dai2019transformer}.
\textbf{HC3 (Human ChatGPT Comparison Corpus)} \cite{hc32023} is the first large-scale human-ChatGPT comparison dataset, containing tens of thousands of comparison responses from human experts and ChatGPT across open-domain, financial, medical, legal, and psychological areas.

Other notable datasets include \textbf{LLM-DetectAIve} \cite{llmdetectaive2024}, which provides 382,330 examples with fine-grained labels distinguishing human-written, fully machine-generated, machine-humanized, and human-polished texts across six domains: arXiv, WikiHow, Wikipedia, Reddit, student essays (OUTFOX), and peer reviews (PeerRead); and \textbf{DetectRL} \cite{wu2025detectrlbenchmarkingllmgeneratedtext}, a real-world benchmark that emphasizes ecological validity in detection scenarios.

Detection methods for AI-generated text can be broadly categorized into zero-shot approaches \cite{mitchell2023detectgptzeroshotmachinegeneratedtext,hans2024spotting}, rewriting-based techniques \cite{mao2024raidargenerativeaidetection,fastdetectgpt2024}, watermarking systems\cite{dathathri2024scalable,zhang2024watermarkinglargelanguagemodels}, and supervised classifiers \cite{roberta_openai,hu2023radar}.

\subsection{Fake News Detection}

Fake news detection has produced quite a few benchmark datasets addressing content veracity.

\textbf{LIAR} \cite{liar2017} dataset contains 12,836 manually labeled short statements collected over a decade from PolitiFact.com. Each statement is labeled with one of six fine-grained truthfulness ratings: pants-fire, false, barely-true, half-true, mostly-true, and true. LIAR-PLUS \cite{liarplus} extends the original LIAR dataset claims with journalist-written justifications and explanations.

\textbf{FakeNewsNet} \cite{shu2019fakenewsnetdatarepositorynews} is a comprehensive data repository containing news content, social context, and spatiotemporal information.
\textbf{FEVER (Fact Extraction and VERification)} \cite{fever2018} contains 185,445 claims manually generated by altering sentences from Wikipedia introductory sections. Each claim is classified as Supported, Refuted, or NotEnoughInfo.  

\textbf{Factify series} released multiple datasets over the previous years. Factify 1 \cite{factify} released a dataset of 50k data points for multi-modal fake news detection. Factify 2 \cite{factify2} provides an additional 50k multimodal data points which also include satire news, to increase the difficulty of the task. Factify5WQA \cite{suresh2024overview} releases 15k data points, paraphrased using AI, and also provides 5W questions to guide detection.

Domain-specific datasets addressing health misinformation include \textbf{CoAID} \cite{coaid2020} and \textbf{CONSTRAINT} \cite{patwa2021fighting}, both targeting COVID-19 misinformation in social media posts, enabling research on real-time misinformation tracking during public health crises.

Other relevant datasets include \textbf{MultiFC} \cite{multifc2019}, \textbf{MMCFND} \cite{mmcfnd_arxiv}, \textbf{Fakeddit} \cite{fakeddit2020} etc.

Fake news detection methods employ diverse approaches including fact-checking \cite{gear2019,Chen_2022,kapalm2023,singhal2024evidence}, multimodal analysis \cite{eann2018,spotfake2020,cafe2022, factify1overview,suryavardan2023findings}, social context integration \cite{defend2019,bigcn2020,gnn_continual2020}, and knowledge graph-based reasoning \cite{kan2021,kgfact2021}.

While existing datasets have advanced both AI-generated text detection and fake news research, significant gaps remain. Most AI-generated text detection datasets rely on synthetic prompts, student essays, or generic web content rather than high-quality journalistic sources. Fake news datasets, though valuable, often focus on short claims or social media posts rather than full-length articles. Furthermore, few datasets systematically compare outputs from multiple state-of-the-art LLMs under controlled conditions. Our dataset aims to address these limitations.

\section{Dataset Creation}

In this section, we describe the process for our dataset creation.

% \begin{figure*}[t!]
%     \centering
%     \subfloat[Huam generated \label{fig:realwc}]{
%         \includegraphics[width=.45\textwidth]{wordcloud_Human_story.png}
%     }\hspace{1em}
%     \subfloat[AI generated\label{fig:realwc} ]{
%         \includegraphics[width=.45\textwidth]{wordcloud_combined_all_llms.png}
%     }\hspace{1em}
%     \caption{Wordcloud of human written and AI generated text. } 
%     \label{fig:wc} 
% \end{figure*}

\subsection{base data source}
Our dataset is built upon a large-scale collection of New York Times (NYT) articles spanning January 1, 2000, to the present day. The core NYT dataset comprises over 2.1 million articles and is updated daily, ensuring that the resource remains current and reflective of evolving news trends. Key metadata fields include the abstract of the article, web URL, headline, keywords, publication date, news desk, section name, byline, word count, among other features. These attributes not only capture the diversity of topics covered by the NYT but also provide rich context for further analysis. %Here is the link to our dataset: 

\begin{figure*}[t!]
    \centering
    
    \subfloat[Human-generated text\label{fig:humanwc}]{
        \includegraphics[width=0.45\textwidth]{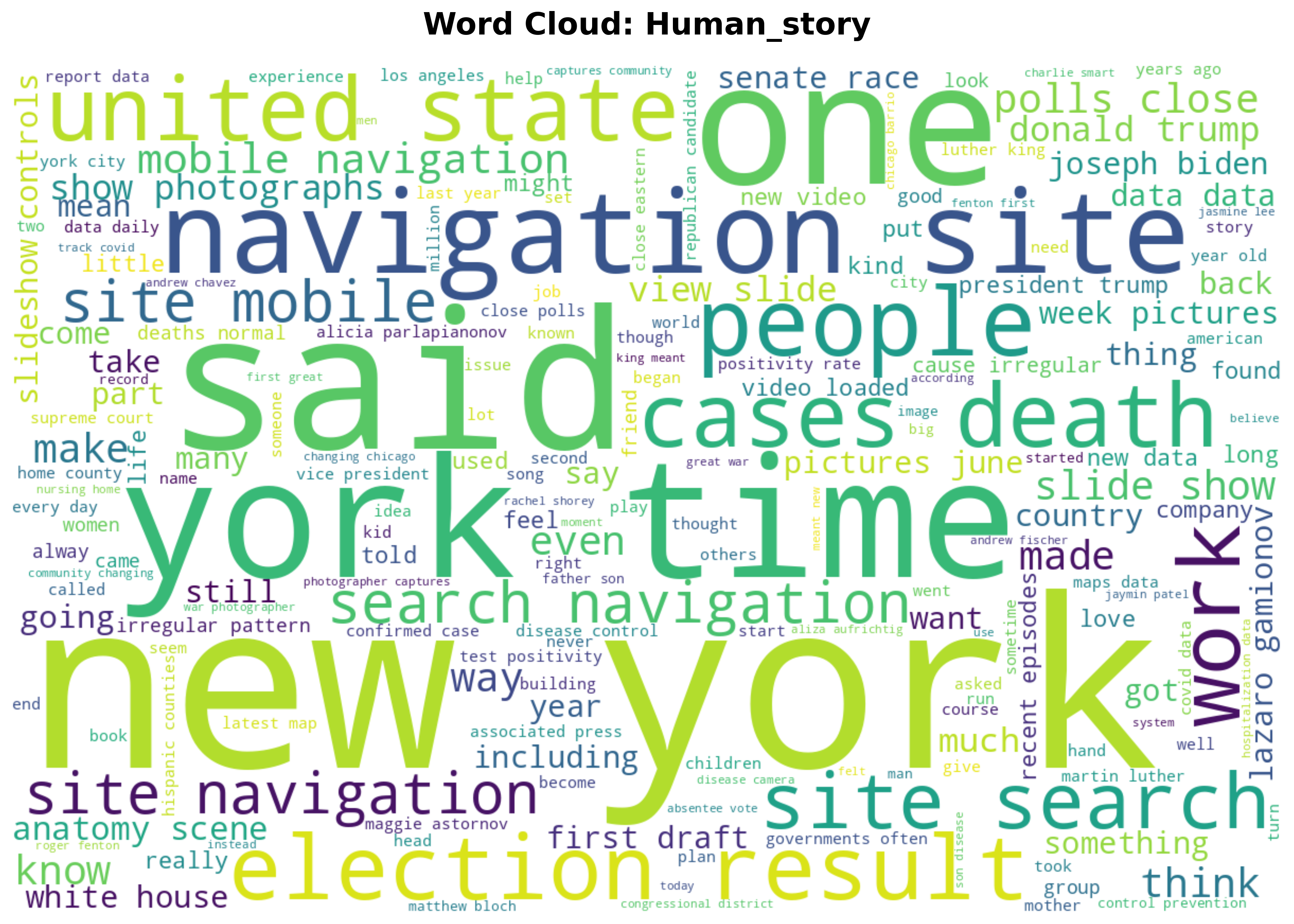}
    }
    \hspace{1em}
    \subfloat[AI-generated text\label{fig:aiwc}]{
        \includegraphics[width=0.45\textwidth]{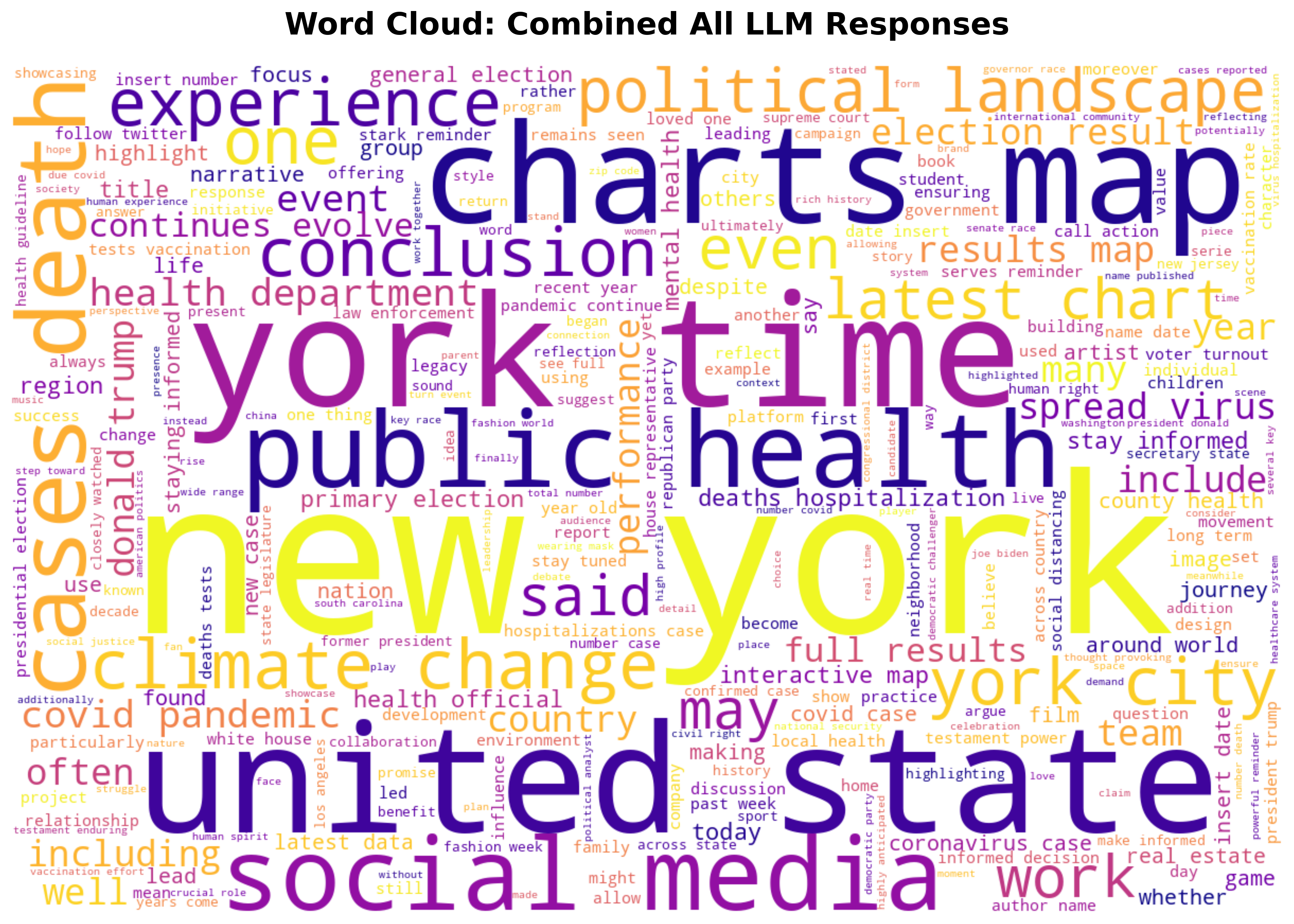}
    }
    
    \caption{Word clouds of human-written and AI-generated text.}
    \label{fig:wc}
\end{figure*}

\subsection{Feature Extraction and Transformation}
\paragraph{Abstract as Prompt:} 
The article abstract is extracted and used as a prompt to generate AI text outputs. This serves as a controlled input for language model experiments, ensuring that each generated output is grounded in real-world journalistic content.

\paragraph{Web URL for Human Story Extraction:} 
The provided web URL for each article is utilized to fetch the full human-authored narrative (hereafter referred to as the ``human story''). This extraction captures the in-depth reporting and storytelling inherent to the source material.

\subsection{Generation of AI Texts}
Using the article abstracts as prompts, several state-of-the-art language models were used to generate corresponding texts. The models include Gemma-2-9B, Mistral-7B, Qwen-2-72B, LLaMA-8B, Yi-Large  and GPT-4o.

The dataset is thus extended to include both the human-authored text (human story) and AI-generated outputs for direct comparison.

\subsection{Dataset Structure and Statistics}
The final assembled dataset is organized into a tabular format with the following columns:

\begin{itemize}
    % \item \texttt{id} --- Unique identifier for each sample.
    % \item \texttt{image} --- The real or model-generated image.
    \item \texttt{text} --- The real or model-generated text.
    \item \texttt{label\_A} --- Binary label indicating whether the text is real (0) or AI-generated (1).
    \item \texttt{label\_B} --- Categorical label indicating the specific generative model (Gemma-2-9B, Mistral-7B, Qwen-2-72B, LLaMA-8B, Yi-Large, or GPT-4o) or human-written text as Human\_story.
\end{itemize}

% \begin{itemize}
%     % \item \textbf{Prompt:} The original abstract extracted from the NYT articles.
%     \item \textbf{Human\_story:} The full narrative text fetched via the article’s web URL.
%     \item \textbf{Gemma-2-9B:} AI-generated text produced using the Gemma-2-9B model.
%     \item \textbf{Mistral-7B:} Output text from the Mistral-7B model.
%     \item \textbf{Qwen-2-72B:} Text generated by the Qwen-2-72B model.
%     \item \textbf{LLaMA-8B:} Output from the LLaMA-8B model.
%     \item \textbf{Yi-Large:} Text generated by the Yi-Large model.
%     \item \textbf{GPT-4o:} AI-generated output from GPT-4o.
% \end{itemize}

% \begin{table}[h]
% \centering
% \begin{tabular}{lr}
% \hline
% \textbf{Column} & \textbf{Count} \\
% \hline
% Prompt         & 7321 \\
% Human\_story   & 7295 \\
% Gemma-2-9B     & 7310 \\
% Mistral-7B     & 7316 \\
% Qwen-2-72B     & 7314 \\
% LLaMA-8B       & 7306 \\
% Yi-Large       & 7319 \\
% GPT-4o       & 7321 \\
% \hline
% \textbf{Total} & 58502 \\
% \hline
% \end{tabular} 
% \caption{Data distribution across real data and varion LLM generated articles. }
% \label{tab:aggregated_counts_total}
% \end{table}

\begin{table*}[h]
\centering
\begin{tabular}{lrrrr}
\hline
\textbf{Source} & \textbf{Train} & \textbf{Validation} & \textbf{Test} & \textbf{Total} \\
\hline
GPT-4o       & 7321 & 1569 & 1569 & 10459 \\
Gemma-2-9B   & 7321 & 1569 & 1568 & 10458 \\
Human\_Story & 7321 & 1569 & 1558 & 10448 \\
LLaMA-8B     & 7321 & 1569 & 1565 & 10455 \\
Mistral-7B   & 7321 & 1569 & 1568 & 10458 \\
Qwen-2-72B   & 7321 & 1569 & 1567 & 10457 \\
Yi-Large     & 7321 & 1569 & 1568 & 10458 \\
\hline
\textbf{Total} & 51247 & 10983 & 10963 & 73193 \\
\hline
\end{tabular}
\caption{Data distribution across real and LLM-generated articles for train, validation, test, and combined splits.}
\label{tab:data_distribution_all_splits}
\end{table*}

In total, we have over 73,000 real and synthetic articles. The detailed distribution is provided in Table \ref{tab:data_distribution_all_splits}. Each LLM contributes roughly equal number of articles. The dataset is available at: https://huggingface.co/datasets/Rajarshi-Roy-research/Defactify\_Text\_Dataset.

Figure \ref{fig:wc} shows the wordclouds of human written and AI generated text (combination of all LLM generated data) respectively. We can see that key words like "new york", "united state" are prominent across both.

\subsection{Potential uses of the dataset}
This dataset is uniquely positioned to advance research in the detection of AI-generated content. With a dual composition of human-authored texts (human stories) and AI-generated outputs from multiple state-of-the-art language models, the dataset provides a rich, annotated resource for developing, training, and benchmarking AI content detection systems. Key research directions include:
\begin{itemize}
    \item \textbf{Develop Robust Classifiers:} Create and evaluate models that distinguish between human-written and AI-generated text, improving the transparency and trustworthiness of digital content.
    \item \textbf{Feature Engineering:} Investigate linguistic, stylistic, and semantic features that are indicative of AI generation. This may involve advanced natural language processing techniques and deep learning methods to capture nuanced patterns that differentiate the two text types.
    \item \textbf{Cross-Model Generalization:} Since the dataset includes outputs from multiple language models (e.g., Gemma-2-9B, Mistral-7B, Qwen-2-72B, LLaMA-8B, Yi-Large, GPT-4o), it offers a comprehensive basis to study whether detectors trained on one model’s output can generalize to others—a crucial aspect given the rapid evolution of AI text generators.
    \item \textbf{Benchmarking and Model Evaluation:} Benchmark the performance of various AI content detection algorithms by comparing metrics such as precision, recall, and robustness under adversarial conditions.
    \item \textbf{Implications for Misinformation and Authenticity:} In an era where AI-generated text can be misused to spread misinformation or manipulate public opinion, robust detection methods are essential for verifying the authenticity of news content. This dataset supports research that aims to mitigate risks associated with the misuse of AI-generated narratives, reinforcing trust in reputable sources like the New York Times.
    \item \textbf{Hybrid Content Recommendation Systems:} Beyond detection, insights derived from this dataset can inform systems that not only recommend content but also flag or annotate texts based on their origin. This hybrid approach would empower users and platforms to navigate mixed content landscapes more effectively.
\end{itemize}
By combining real-world, human-authored journalism with a spectrum of AI-generated texts, our dataset offers an unprecedented resource for exploring and advancing AI content detection methodologies, ultimately contributing to more secure and transparent information ecosystems.

\subsection{Baseline}

We implement a classification-based baseline inspired by the Raidar method \cite{mao2024raidargenerativeaidetection}, which detects machine-generated content via rewriting. The key idea is that large language models (LLMs) tend to make fewer edits when rewriting content originally generated by another LLM, as opposed to human-written text, which they tend to revise more extensively. This concept can be further extended to hypothesize that an LLM will make even fewer edits when rewriting text originally generated by itself, compared to text generated by a different LLM. This is due to the structural and stylistic consistencies within the outputs of a given LLM, which make its own generations more "acceptable" or lower-loss candidates during rewriting. As a result, measuring the degree of rewriting not only helps distinguish AI-generated from human text, but also allows us to infer the specific model that likely generated the input.

As illustrated in Figure~\ref{fig:baseline_pipeline}, our method prompts a fixed rewriting model (e.g., GPT-3.5-Turbo) with a standardized instruction (e.g., “Concise this for me and keep all the information”) to rewrite the input text. The figure visually demonstrates that human-written inputs typically result in larger modifications during rewriting—highlighted by numerous character-level insertions and deletions—whereas AI-generated inputs lead to much smaller changes, reflecting the model's tendency to preserve its own generative style.

In our classification-based baseline, for each input text, we prompt multiple candidate LLMs (e.g., GPT variants) to rewrite the text and compute the edit distance between their rewritten output and the original input using the Levenshtein metric. The model whose rewrite is closest (i.e., has the minimum edit distance) is predicted as the most likely generator of the input. If all edit distances exceed a predefined threshold—chosen as the \textbf{median} of the \textbf{maximum edit distance} across training samples—the input is classified as human-written. 

\begin{center}
    \includegraphics[width=0.9\textwidth]{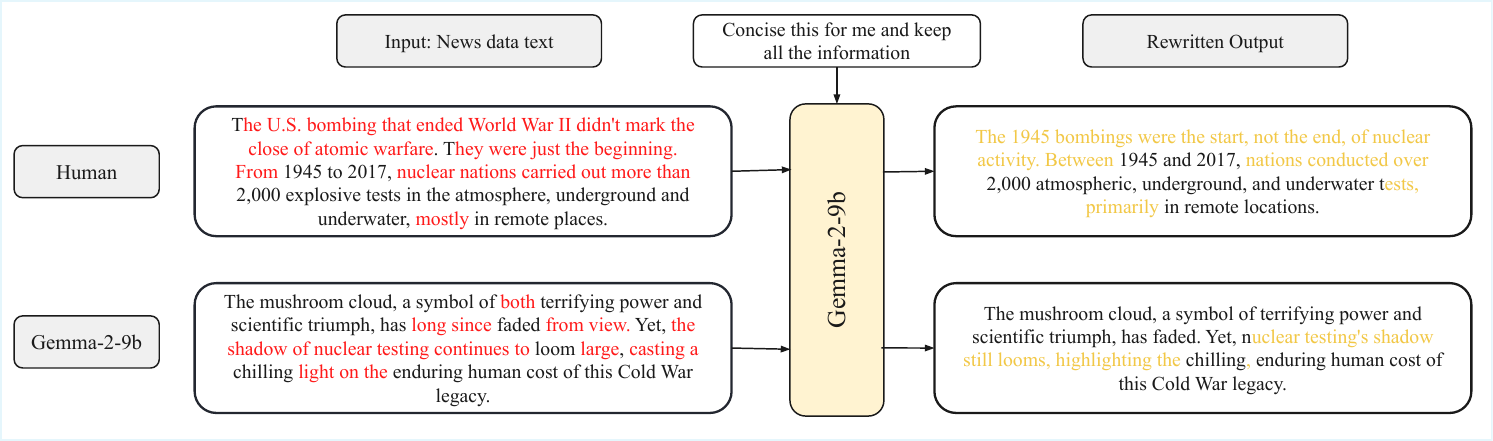}
    \captionof{figure}{\textbf{Illustration of Raidar concept.} Given a News data text and an LLM-generated text, the same LLM is asked to rewrite the inputs while preserving meaning. The rewriting of a human-written text undergoes more character-level edits (highlighted in red/yellow), while the rewriting of an LLM-written text remains largely unchanged.}
    \label{fig:baseline_pipeline}
\end{center}

\section{Results}

We define 2 tasks on the dataset: Human vs AI generated text classification (task A) and model attribution for AI-generated text (task B). The results of our baseline on both the tasks are provided in table \ref{tab:result}. These results establish foundational performance metrics for two key classification tasks based on our dataset, and serve as reference points for future research and method development.

\begin{table}[h]
\centering
\begin{tabular}{|l|c|c|}
\hline
\textbf{Task} & \textbf{Description} & \textbf{Accuracy} \\
\hline
Task A & Human vs. AI-generated text classification & 0.5300 \\
\hline
Task B & Model attribution for AI-generated text    & 0.0504 \\
\hline
\end{tabular}
\caption{Results of our baseline. }
\label{tab:result}
\end{table}

For Task A, the baseline uses an edit-distance rewriting approach inspired by Raidar~\cite{mao2024raidargenerativeaidetection} to distinguish human-written from AI-generated texts, achieving an accuracy of 53\%. For Task B, the same methodology is employed to identify the originating model of AI-generated texts, yielding an accuracy of 5.04\%.

These baseline scores demonstrate the inherent difficulty of both tasks, especially as modern language models continue to improve. The dataset enables the development and benchmarking of more sophisticated detection and attribution techniques.

\section{Conclusion}
In this paper. We describe and release a comprehensive dataset containing over 73,000 human written and AI generated news samples. Six LLMs are used for AI generated content. We propose two tasks based on the dataset : AI generated text detection and model attribution for AI generated text. 

By bridging real-world journalism with modern LLM-generated content, our dataset establishes a rigorous benchmark for advancing AI-generated text detection in domains where content authenticity and journalistic integrity are important. The low baseline results underscore the difficulty of these tasks and the substantial room for methodological innovation in this critical research area.

Future research could include expanding the dataset to other languages and modalities. LLM generated news can potentially  be made more human-like challenging via in-context learning \cite{patwa2024enhancing}, or finetuning \cite{kaddour2024syntheticdatagenerationlowresource}  to teach LLMs to generate more human-like text. More sophisticated methods to detect AI generated text is also an interesting line of future work. 

% \printbibliography
% \bibliographystyle{plainnat} % or another style like abbrvnat, unsrtnat
\bibliography{references} % Ensure you have a references.bib file
\end{document}